\documentclass[10pt, conference, compsocconf]{IEEEtran}
\usepackage{times}
\usepackage{helvet}
\usepackage{courier}
\usepackage[hyphens]{url}
\usepackage{graphicx}
\usepackage{balance}
\usepackage{color}
\usepackage{dblfloatfix}

\title{Solving Arithmetic Word Problems Automatically Using Transformer and Unambiguous Representations}

\author{\IEEEauthorblockN{Kaden Griffith and Jugal Kalita}
\IEEEauthorblockA{University of Colorado Colorado Springs \\
Colorado Springs, Colorado \\
Email: kadengriffith@gmail.com and jkalita@uccs.edu}
}

\begin{document}
    \maketitle

    \begin{abstract}
        Constructing accurate and automatic solvers of math word problems has proven to be quite challenging.
        Prior attempts using machine learning have been trained on corpora specific to math word problems to produce arithmetic expressions in infix notation before answer computation.
        We find that custom-built neural networks have struggled to generalize well.
        This paper outlines the use of Transformer networks trained to translate math word problems to equivalent arithmetic expressions in infix, prefix, and postfix notations.
        In addition to training directly on domain-specific corpora, we use an approach that pre-trains on a general text corpus to provide foundational language abilities to explore if it improves performance.
        We compare results produced by a large number of neural configurations and find that most configurations outperform previously reported approaches on three of four datasets with significant increases in accuracy of over 20 percentage points.
        The best neural approaches boost accuracy by almost 10\% on average when compared to the previous state of the art.
    \end{abstract}


    \section{Introduction}
        Students are exposed to simple arithmetic word problems starting in elementary school, and most become proficient in solving them at a young age.
        Automatic solvers of such problems could potentially help educators, as well as become an integral part of general question answering services.
        However, it has been challenging to write programs to solve even such elementary school level problems well.

        Solving a math word problem (MWP) starts with one or more sentences describing a transactional situation to be understood.
        The sentences are processed to produce an arithmetic expression, which is evaluated to provide an answer.
        Recent neural approaches to solving arithmetic word problems have used various flavors of recurrent neural networks (RNN) as well as reinforcement learning.
        Such methods have had difficulty achieving a high level of generalization.
        Often, systems extract the relevant numbers successfully but misplace them in the generated expressions.
        More problematic, they get the arithmetic operations wrong.
        The use of infix notation also requires pairs of parentheses to be placed and balanced correctly, bracketing the right numbers.
        There have been problems with parentheses placement as well.
        \begin{figure}[!ht]
            \caption{Possible generated expressions for a MWP.}
            \label{figure:misrepresentations}
            \centering
            \begin{tabular}{p{0.9\linewidth}}
            \hline
            { \bf Question: } \\
            At the fair Adam bought 13 tickets. After riding the ferris wheel he had 4 tickets left. If each ticket cost 9 dollars, how much money did Adam spend riding the ferris wheel? \\ [.05in]
            { \bf Some possible expressions that can be produced: } \\
            {\small $(13 - 4) * 9, 9 * 13 - 4, 5 * 13 - 4, 13 - 4 * 9, 13 - (4 * 9),$ } \\
            {\small $(9 * 13 - 4), (9) * 13 -4, (9) * 13 - (4)$, etc. } \\
            \hline
            \end{tabular}
        \end{figure}
        Correctly extracting the numbers in the problem is necessary.  Figure \ref{figure:misrepresentations} gives examples of some infix representations that a machine learning  solver can potentially produce from a simple word problem using the correct numbers.
        Of the expressions shown, only the first one is correct.
        After carefully observing expressions that actual problem solvers have generated, we want to explore if the use of infix notation may itself be a part of the problem because it requires the generation of additional characters, the open and close parentheses, which must be balanced and placed correctly.

        The actual numbers appearing in MWPs vary widely from problem to problem.
        Real numbers take any conceivable value, making it almost impossible for a neural network to learn representations for them.
        As a result, trained programs sometimes generate expressions that have seemingly random numbers.
        For example, in some runs, a trained program could generate a potentially inexplicable expression such as $(25.01 - 4) * 9$ for the problem given in Figure \ref{figure:misrepresentations}, with one or more numbers not in the problem sentences.
        We hypothesize that replacing the numbers in the problem statement with generic tags like $\rm \langle n1 \rangle$, $\rm \langle n2 \rangle$, and $\rm \langle n3 \rangle$ and saving their values as a pre-processing step, does not take away from the generality of the solution, but suppresses the problem of fertility in number generation leading to the introduction of numbers not present in the question sentences.

        Another idea we want to test is whether a neural network which has been pre-trained to acquire language knowledge is better able to ``understand" the problem sentences.
        Pre-training with a large amount of arithmetic-related text is likely to help develop such knowledge, but due to the lack of large such focused corpora, we want to test whether pre-training with a sufficient general corpus is beneficial.

        In this paper, we use the Transformer model \cite{vaswani2017attention} to solve arithmetic word problems as a particular case of machine translation from text to the language of arithmetic expressions.
        Transformers in various configurations have become a staple of NLP in the past two years.
        Past neural approaches did not treat this problem as pure translation like we do, and additionally, these approaches usually augmented the neural architectures with various external modules such as parse trees or used deep reinforcement learning, which we do not do.
        In this paper, we demonstrate that Transformers can be used to solve MWPs successfully with the simple adjustments we describe above.
        We compare performance on four individual datasets.
        In particular, we show that our translation-based approach outperforms state-of-the-art results reported by \cite{wang2018mathdqn,hosseini2014learning,kushman2014learning,roy2015reasoning,robaidek2018data} by a large margin on three of four datasets tested.
        On average, our best neural architecture outperforms previous results by almost 10\%, although our approach is conceptually more straightforward.

        We organize our paper as follows.
        The second section presents related work.
        Then, we discuss our approach.
        We follow by an analysis of experimental results and compare them to those of other recent approaches.
        We also discuss our successes and shortcomings.
        Finally, we share our concluding thoughts and end with our direction for future work.

    \section{Related Work}
        Past strategies have used rules and templates to match sentences to arithmetic expressions.
        Some such approaches seemed to solve problems impressively within a narrow domain, but performed poorly when out of domain, lacking generality  \cite{bobrow1964natural,bakman2007robust,liguda2012modeling,shi2015automatically}.
        Kushman et al. \cite{kushman2014learning} used feature extraction and template-based categorization by representing equations as expression forests and finding a near match.
        Such methods required human intervention in the form of feature engineering and development of templates and rules, which is not desirable for expandability and adaptability.
        Hosseini et al. \cite{hosseini2014learning} performed statistical similarity analysis to obtain acceptable results, but did not perform well with texts that were dissimilar to training examples.

        Existing approaches have used various forms of auxiliary information.
        Hosseini et al. \cite{hosseini2014learning} used verb categorization to identify important mathematical cues and contexts.
        Mitra and Baral \cite{mitra2016learning} used predefined formulas to assist in matching.
        Koncel-Kedziorski et al. \cite{koncel2015parsing} parsed the input sentences, enumerated all parses, and learned to match, requiring expensive computations.
        Roy and Roth \cite{roy2017unit} performed searches for semantic trees over large spaces.

        Some recent approaches have transitioned to using neural networks.
        Semantic parsing takes advantage of RNN architectures to parse MWPs directly into equations or expressions in a  math-specific language \cite{shi2015automatically,sun2019neural}.
        RNNs have shown promising results, but they have had difficulties balancing parenthesis, and also, sometimes incorrectly choose numbers when generating equations.
        Rehman et al. \cite{rehman2019automatically} used  POS tagging and classification of equation templates to produce systems of equations from third-grade level MWPs.
        Most recently, Sun et al. \cite{sun2019neural} used a Bi-Directional LSTM architecture for math word problems.
        Huang et al. \cite{huang-etal-2018-neural} used a deep reinforcement learning model to achieve character placement in both seen and novel equation templates.
        Wang et al. \cite{wang2018mathdqn} also used deep reinforcement learning.

    \section{Approach}
        We view math word problem solving as a sequence-to-sequence translation problem.
        RNNs have excelled in sequence-to-sequence problems such as translation and question answering.
        The recent introduction of attention mechanisms has improved the performance of RNN models.
        Vaswani et al.  \cite{vaswani2017attention} introduced the Transformer network, which uses stacks of attention layers instead of recurrence.
        Applications of Transformers have achieved state-of-the-art performance in many NLP tasks.
        We use this architecture to produce character sequences that are arithmetic expressions.
        The models we experiment with are easy and efficient to train, allowing us to test several configurations for a comprehensive comparison.
        We use several configurations of Transformer networks to learn the prefix, postfix, and infix notations of MWP equations independently.

        Prefix and postfix representations of equations do not contain parentheses, which has been a source of confusion in some approaches.
        If the learned target sequences are simple, with fewer characters to generate, it is less likely to make mistakes during generation.
        Simple targets also may help the learning of the model to be more robust.
        Experimenting with all three representations for equivalent expressions may help us discover which one works best.

        We train on standard datasets, which are readily available and commonly used.
        Our method considers the translation of English text to simple algebraic expressions.
        After performing experiments by training directly on math word problem corpora, we perform a different set of experiments by pre-training on a  general language corpus.
        The success of pre-trained models such as ELMo \cite{peters2018deep}, GPT-2 \cite{budzianowski2019hello}, and BERT \cite{devlin2018bert} for many natural language tasks, provides reasoning that pre-training is likely to produce better learning by our system.
        We use pre-training so that the system has some foundational knowledge of English before we train it on the domain-specific text of math word problems.
        However, the output is not natural language but algebraic expressions, which is likely to limit the effectiveness of such pre-training.

    \subsection{Data}
        We work with four individual datasets.
        The datasets contain addition, subtraction, multiplication, and division word problems.
        \begin{enumerate}
            \item {\bf AI2} \cite{hosseini2014learning}.
            AI2 is a collection of 395 addition and subtraction problems, containing numeric values, where some may not be relevant to the question.
            \item {\bf CC} \cite{roy2016solving}.
            The Common Core dataset contains 600 2-step  questions.
            The Cognitive Computation Group at the University of Pennsylvania\footnote{\url{https://cogcomp.seas.upenn.edu/page/demos/}} gathered these questions.
            \item {\bf IL} \cite{roy2015reasoning}.
            The Illinois dataset contains 562 1-step algebra word questions.
            The Cognitive Computation Group compiled these questions also.
            \item {\bf MAWPS} \cite{koncel2016mawps}.
            MAWPS is a relatively large collection, primarily from other MWP datasets.
            We use 2,373 of 3,915 MWPs from this set.
            The problems not used 
            were more complex problems that generate systems of equations.
            We exclude such problems because generating systems of equations is not our focus.
        \end{enumerate}

        We take a randomly sampled 95\% of examples from each dataset for training.
        From each dataset, MWPs not included in training make up the testing data used when generating our results.
        Training and testing are repeated three times, and reported results are an average of the three outcomes.

    \subsection{Representation Conversion}
        We take a simple approach to convert infix expressions found in the MWPs to the other two representations.
        Two stacks are filled by iterating through string characters, one with operators found in the equation and the other with the operands.
        From these stacks, we form a binary tree structure.
        Traversing an expression tree in pre-order results in a prefix conversion.
        Post-order traversal gives us a postfix expression.
        Three versions of our training and testing data are created to correspond to each type of expression.
        By training on different representations, we expect our test results to change.

    \subsection{Pre-training}
        We pre-train half of our networks to endow them with a foundational knowledge of English.
        Pre-training models on significant-sized language corpora have been a common approach recently.
        We explore the pre-training approach using a general English corpus because the language of MWPs is regular English, interspersed with numerical values.
        Ideally, the corpus for pre-training should be a very general and comprehensive corpus like an English Wikipedia dump or many gigabytes of human-generated text scraped from the internet like GPT-2 \cite{radford2019language} used.
        However, in this paper, we want to perform experiments to see if pre-training with a smaller corpus can help.
        In particular, for this task, we use the IMDb Movie Reviews dataset \cite{maas-EtAl:2011:ACL-HLT2011}.
        This set contains 314,041 unique sentences.
        Since movie reviewers wrote this data, it is a reference to natural language not related to arithmetic.
        Training on a much bigger and general corpus may make the language model stronger, but we leave this for future work.

        We compare pre-trained models to non-pre-trained models to observe performance differences.
        Our pre-trained models are trained in an unsupervised fashion to improve the encodings of our fine-tuned solvers.
        In the pre-training process, we use sentences from the IMDb reviews with a target output of an empty string.
        We leave the input unlabelled, which focuses the network on adjusting encodings while providing unbiased decoding when we later change from IMDb English text to MWP-Data.

    \subsection{Method: Training and Testing}
        The input sequence is a natural language specification of an arithmetic word problem.
        The MWP questions and equations have been encoded using the subword text encoder provided by the TensorFlow Datasets library.
        The output is an expression in prefix, infix, or postfix notation, which then can be manipulated further and solved to obtain a final answer.

        All examples in the datasets contain numbers, some of which are unique or rare in the corpus.
        Rare terms are adverse for generalization since the network is unlikely to form good representations for them.
        As a remedy to this issue, our networks do not consider any relevant numbers during training.
        Before the networks attempt any translation, we pre-process each question and expression by a number mapping algorithm.
        This algorithm replaces each numeric value with a corresponding identifier (e.g., $\langle n1 \rangle$, $\langle n2 \rangle$, etc.), and remembers the necessary mapping.
        We expect that this approach may significantly improve how networks interpret each question.
        When translating, the numbers in the original question are tagged and cached.
        From the encoded English and tags, a predicted sequence resembling an expression presents itself as output.
        Since each network's learned output resembles an arithmetic expression (e.g., $\langle n1 \rangle + \langle n2 \rangle * \langle n3 \rangle$), we use the cached tag mapping to replace the tags with the corresponding numbers and return a final mathematical expression.

        Three representation models are trained and tested separately: Prefix-Transformer, Postfix-Transformer, and Infix-Transformer.
        For each experiment, we use representation-specific Transformer architectures.
        Each model uses the Adam optimizer with $beta_1=0.95$ and $beta_2=0.99$ with a standard epsilon of $1 \times e^{-9}$.
        The learning rate is reduced automatically in each training session as the loss decreases.
        Throughout the training, each model respects a 10\% dropout rate.
        We employ a batch size of 128 for all training.
        Each model is trained on MWP data for 300 iterations before testing.
        The networks are trained on a machine using 1 Nvidia 1080 Ti graphics processing unit (GPU).

        We compare medium-sized, small, and minimal networks to show if network size can be reduced to increase training and testing efficiency while retaining high accuracy.
        Networks over six layers have shown to be non-effective for this task.
        We tried many configurations of our network models, but report results with only three configurations of Transformers.

        \begin{enumerate}
            \item[-] {\bf Transformer Type 1:}
            This network is a small to medium-sized network consisting of 4 Transformer layers.
            Each layer utilizes 8 attention heads with a depth of 512 and a feed-forward depth of 1024.
            \item[-] {\bf Transformer Type 2:}
            The second model is small in size, using 2 Transformer layers.
            The layers utilize 8 attention heads with a depth of 256 and a feed-forward depth of 1024.
            \item[-] {\bf Transformer Type 3:}
            The third type of model is minimal, using only 1 Transformer layer.
            This network utilizes 8 attention heads with a depth of 256 and a feed-forward depth of 512.
        \end{enumerate}

        \paragraph{Objective Function}
        We calculate the loss in training according to a mean of the sparse categorical cross-entropy formula.
        Sparse categorical cross-entropy \cite{de2005tutorial} is used for identifying classes from a feature set, which assumes a large target classification set.
        Evaluation between the possible translation classes (all vocabulary subword tokens) and the produced class (predicted token) is the metric of performance here.
        During each evaluation, target terms are masked, predicted, and then compared to the masked (known) value.
        We adjust the model's loss according to the mean of the translation accuracy after predicting every determined subword in a translation.
        \begin{equation}
            \label{eq:loss}
            loss = \displaystyle\sum_{i=1}^{I} \frac{1}{J}\displaystyle\sum_{j=1}^{J} \bigg(-\displaystyle\sum_{k=1}^{K} target_{j,k}*log \big(p(j \in k)\big) \bigg)
        \end{equation}
        where
        $K = |Translation \; Classes|$,
        $J = |Translation|$, and
        $I$ is the number of examples.

        \paragraph{Experiment 1: Representation}
        Some of the problems encountered by prior approaches seem to be attributable to the use of infix notation.
        In this experiment, we compare translation BLEU-2 scores to spot the differences in representation interpretability.
        Traditionally, a BLEU score is a metric of translation quality \cite{papineni2002bleu}.
        Our presented BLEU scores represent an average of scores a given model received over each of the target test sets.
        We use a standard bi-gram weight to show how accurate translations are within a window of two adjacent terms.
        After testing translations, we calculate an average BLEU-2 score per test set, which is related to the success over that data.
        An average of the scores for each dataset become the presented value.
        \begin{equation}
            \label{eq:averageBLEU}
            model_{avg} = \frac{1}{N}\displaystyle\sum_{n=1}^{N} BLEUavg_{n}
        \end{equation}
        where $N$ is the number of test datasets, which is 4.

        \paragraph{Experiment 2: State-of-the-art}
        This experiment compares our networks to recent previous work.
        We count a given test score by a simple ``correct versus incorrect" method.
        The answer to an expression directly ties to all of the translation terms being correct, which is why we do not consider partial precision.
        We compare average accuracies over 3 test trials on different randomly sampled test sets from each MWP dataset.
        This calculation more accurately depicts the generalization of our networks.

        \paragraph{Effect of Pre-training}
        We also explore the effect of language pre-training, as discussed earlier.
        This training occurs over 30 iterations, at the start of the two experiments, to introduce a good level of language understanding before training on the MWP data.
        The same Transformer architectures are also trained solely on the MWP data.
        We calculate the reported results as:
        \begin{equation}
            \label{eq:averageAccuracy}
            model_{avg} = \frac{1}{R}\displaystyle\sum_{r=1}^{R} \bigg( \frac{1}{N}\displaystyle\sum_{n=1}^{N} \frac{C \in n}{P \in n} \bigg)
        \end{equation}
        where $R$ is the number of test repetitions, which is 3;
        $N$ is the number of test datasets, which is 4;
        $P$ is the number of MWPs, and
        $C$ is the number of correct equation translations.

  \section{Results}
    We now present the results of our various experiments.
    We compare the three representations of target equations and three architectures of the Transformer model in each test.

    Results of Experiment 1 are given in Table \ref{table:Experiment1BLEU}.
    For clarity, the number in parentheses in front of a row is the Transformer type.
    By using BLEU scores, we assess the translation capability of each network.
    This test displays how networks transform different math representations to a character summary level.
    \begin{table}[!ht]
        \caption{BLEU-2 comparison for Experiment 1.}
        \label{table:Experiment1BLEU}
        \begin{center}
        \begin{tabular}{p{0.6\linewidth}p{0.29\linewidth}}
        \hline
        {\small \bf (Type) Model } & {\small \bf Average } \\
        \hline
        {\small \em Pre-trained } & \\
        {\small (1) Prefix-Transformer } & {\small 94.03 } \\
        {\small (1) Postfix-Transformer } & {\small 92.61 } \\
        {\small (1) Infix-Transformer } & {\small 86.24 } \\
        {\small (2) Prefix-Transformer } & {\small 93.51 } \\
        {\small (2) Postfix-Transformer } & {\small 92.88 } \\
        {\small (2) Infix-Transformer } & {\small 87.14 } \\
        {\small (3) Prefix-Transformer } & {\small 93.39 } \\
        {\small (3) Postfix-Transformer } & {\small 93.03 } \\
        {\small (3) Infix-Transformer } & {\small 86.72 } \\
        {\small \em Non-pre-trained } & \\
        {\small (1) Prefix-Transformer } & {\small 94.95 } \\
        {\small (1) Postfix-Transformer } & {\small 87.26 } \\
        {\small (1) Infix-Transformer } & {\small 87.86 } \\
        {\small (2) Prefix-Transformer } & {\small \bf 95.57 } \\
        {\small (2) Postfix-Transformer } & {\small 94.28 } \\
        {\small (2) Infix-Transformer } & {\small 89.16 } \\
        {\small (3) Prefix-Transformer } & {\small 95.13 } \\
        {\small (3) Postfix-Transformer } & {\small 94.17 } \\
        {\small (3) Infix-Transformer } & {\small 89.22 } \\
        \hline
        \end{tabular}
        \end{center}
    \end{table}
    \begin{table}[!ht]
        \caption{Summary of BLEU scores from Table \ref{table:Experiment1BLEU}.}
        \label{table:Experiment1BLEUAverages}
        \begin{center}
        \begin{tabular}{p{0.6\linewidth}p{0.29\linewidth}}
        \hline
        {\small \bf Description} & {\small \bf Average} \\
        \hline
        {\small All models } & {\small 91.51 } \\
        {\small All prefix models } & {\small \bf 94.43 } \\
        {\small All postfix models } & {\small 92.37 } \\
        {\small All infix models } & {\small 87.72 } \\
        {\small All pre-trained models } & {\small 91.06 } \\
        {\small All non-pre-trained models } & {\small 91.96 } \\
        {\small All medium (type 1) models } & {\small 90.49 } \\
        {\small All small (type 2) models } & {\small 92.09 } \\
        {\small All minimal (type 3) models } & {\small 91.94 } \\
        \hline
        \end{tabular}
        \end{center}
    \end{table}

    We compare by average BLEU-2 accuracy among our tests in the {\em Average} column of Table \ref{table:Experiment1BLEU} to communicate these translation differences.
    To make it easier to understand the results, Table \ref{table:Experiment1BLEUAverages} provides a summary of Table \ref{table:Experiment1BLEU}.

    Looking at Tables \ref{table:Experiment1BLEU} and \ref{table:Experiment1BLEUAverages}, we note that both the prefix and postfix representations of our target language perform better than the generally used infix notation.
    The non-pre-trained models perform slightly better than the pre-trained models, and the small or Type 2 models perform slightly better than the minimal-sized and medium-sized Transformer models.
    The non-pre-trained type 2 prefix Transformer arrangement produced the most consistent translations.

    \begin{table*}[!ht]
        \caption{Test results for Experiment 2 (* denotes averages on present values only). }
        \label{table:Experiment2Results}
        \centering
        \begin{tabular}{p{0.29\linewidth}p{0.11\linewidth}p{0.11\linewidth}p{0.11\linewidth}p{0.13\linewidth}p{0.1\linewidth}}
        \hline
        {\small \bf (Type) Model } & {\small \bf AI2 } & {\small \bf CC } & {\small \bf IL } & {\small \bf MAWPS } & {\small \bf Average } \\
        \hline
        {\small \cite{hosseini2014learning} Hosseini, et.al. } & {\small 77.7 } & {\small -- } & {\small -- } & {\small -- } & {\small $^*$77.7 } \\
        {\small \cite{kushman2014learning} Kushman, et.al. } & {\small 64.0 } & {\small 73.7 } & {\small 2.3 } & {\small -- } & {\small $^*$46.7 } \\
        {\small \cite{roy2015reasoning} Roy, et.al. } & {\small -- } & {\small -- } & {\small 52.7 } & {\small -- } & {\small $^*$52.7 } \\
        {\small \cite{robaidek2018data} Robaidek, et.al. } & {\small -- } & {\small -- } & {\small -- } & {\small 62.8 } & {\small $^*$62.8 } \\
        {\small \cite{wang2018mathdqn} Wang, et.al. } & {\small \bf 78.5 } & {\small 75.5 } & {\small 73.3 } & {\small -- } & {\small $^*$75.4 } \\
        {\small \em Pre-trained } & & & & & \\
        {\small (1) Prefix-Transformer } & {\small 70.2 } & {\small 91.1 } & {\small 95.2 } & {\small 82.4 } & {\small 84.7 } \\
        {\small (1) Postfix-Transformer } & {\small 68.4 } & {\small 90.0 } & {\small 92.9 } & {\small 82.7 } & {\small 83.5 } \\
        {\small (1) Infix-Transformer } & {\small 75.4 } & {\small 74.4 } & {\small 64.3 } & {\small 56.4 } & {\small 67.6 } \\
        {\small (2) Prefix-Transformer } & {\small 66.7 } & {\small 91.1 } & {\small \bf 96.4 } & {\small 82.1 } & {\small 84.1 } \\
        {\small (2) Postfix-Transformer } & {\small 73.7 } & {\small 93.3 } & {\small 94.1 } & {\small 82.4 } & {\small 85.9 } \\
        {\small (2) Infix-Transformer } & {\small 75.4 } & {\small 75.6 } & {\small 66.7 } & {\small 59.0 } & {\small 69.2 } \\
        {\small (3) Prefix-Transformer } & {\small 70.2 } & {\small 91.1 } & {\small 95.2 } & {\small 82.4 } & {\small 84.7 } \\
        {\small (3) Postfix-Transformer } & {\small 73.7 } & {\small 92.2 } & {\small 94.1 } & {\small 82.1 } & {\small 85.5 } \\
        {\small (3) Infix-Transformer } & {\small 75.4 } & {\small 75.6 } & {\small 64.3 } & {\small 58.7 } & {\small 68.5 } \\
        {\small \em Non-pre-trained } & & & & & \\
        {\small (1) Prefix-Transformer } & {\small 71.9 } & {\small \bf 94.4 } & {\small 95.2 } & {\small 83.4 } & {\small 86.3 } \\
        {\small (1) Postfix-Transformer } & {\small 73.7 } & {\small 81.1 } & {\small 92.9 } & {\small 75.7 } & {\small 80.8 } \\
        {\small (1) Infix-Transformer } & {\small 77.2 } & {\small 73.3 } & {\small 61.9 } & {\small 56.8 } & {\small 67.3 } \\
        {\small (2) Prefix-Transformer } & {\small 71.9 } & {\small \bf 94.4 } & {\small 94.1 } & {\small \bf 84.7 } & {\small 86.3 } \\
        {\small (2) Postfix-Transformer } & {\small 77.2 } & {\small \bf 94.4 } & {\small 94.1 } & {\small 83.1 } & {\small \bf 87.2 } \\
        {\small (2) Infix-Transformer } & {\small 77.2 } & {\small 76.7 } & {\small 66.7 } & {\small 61.5 } & {\small 70.5 } \\
        {\small (3) Prefix-Transformer } & {\small 71.9 } & {\small 93.3 } & {\small 95.2 } & {\small 84.1 } & {\small 86.2 } \\
        {\small (3) Postfix-Transformer } & {\small 77.2 } & {\small 94.4 } & {\small 94.1 } & {\small 82.4 } & {\small 87.0 } \\
        {\small (3) Infix-Transformer } & {\small 77.2 } & {\small 76.7 } & {\small 66.7 } & {\small 62.4 } & {\small 70.7 } \\
        \hline
        \end{tabular}
    \end{table*}
    \begin{table}[!ht]
        \caption{Summary of Accuracies from Table \ref{table:Experiment2Results}.}
        \label{table:Experiment2ResultAverages}
        \begin{center}
        \begin{tabular}{p{0.6\linewidth}p{0.29\linewidth}}
        \hline
        {\small \bf Description} & {\small \bf Average} \\
        \hline
        {\small All models } & {\small 79.78 } \\
        {\small All prefix models } & {\small \bf 85.37 } \\
        {\small All postfix models } & {\small 84.99 } \\
        {\small All infix models } & {\small 68.97 } \\
        {\small All pre-trained models } & {\small 79.30 } \\
        {\small All non-pre-trained models } & {\small 80.25 } \\
        {\small All medium (type 1) models } & {\small 78.38 } \\
        {\small All small (type 2) models } & {\small 80.51 } \\
        {\small All minimal (type 3) models } & {\small 80.44 } \\
        \hline
        \end{tabular}
        \end{center}
    \end{table}

    Table \ref{table:Experiment2Results} provides detailed results of Experiment 2.
    The numbers are absolute accuracies, i.e., they correspond to cases where the arithmetic expression generated is 100\% correct, leading to the correct numeric answer.
    Results by \cite{wang2018mathdqn,hosseini2014learning,roy2015reasoning,robaidek2018data} are sparse but indicate the scale of success compared to recent past approaches.
    Prefix, postfix, and infix representations in Table \ref{table:Experiment2Results} show that network capabilities are changed by how teachable the target data is.
    The values in the last column of Table \ref{table:Experiment2Results} are summarized in Table \ref{table:Experiment2ResultAverages}.
    How the models compare with respect to accuracy closely resembles the comparison of BLEU scores, presented earlier.
    Thus, BLEU scores seem to correlate well with accuracy values in our case.

    While our networks fell short of \cite{wang2018mathdqn} AI2 testing accuracy, we present state-of-the-art results for the remaining three datasets.
    The AI2 dataset is tricky because it has numeric values in the word descriptions that are extraneous or irrelevant to the actual computation, whereas the other datasets have only relevant numeric values.
    The type 2 postfix Transformer received the highest testing average of 87.2\%.

    Our attempt at language pre-training fell short of our expectations in all but one tested dataset.
    We had hoped that more stable language understanding would improve results in general.
    As previously mentioned, using more general and comprehensive corpora of language could help grow semantic ability.

    \subsection{Analysis}
        All of the network configurations used were very successful for our task.
        The prefix representation overall provides the most stable network performance.
        To display the capability of our most successful model (type 2 postfix Transformer), we present some outputs of the network in Figure \ref{figure:successfulTranslations}.

        The models respect the syntax of math expressions, even when incorrect.
        For the majority of questions, our translators were able to determine operators based solely on the context of language.
        \begin{figure}
            \caption{Successful postfix translations.}
            \label{figure:successfulTranslations}
            \centering
            \begin{tabular}{p{0.9\linewidth}}
            \hline
            {\small \bf AI2 } \\
            {\small A spaceship traveled 0.5 light-year from earth to planet x and 0.1 light-year from planet x to planet y. Then it traveled 0.1 light-year from planet y back to Earth. How many light-years did the spaceship travel in all? } \\ [.05in]
            {\small \em Translation produced: } \\
            {\small 0.5 0.1 + 0.1 + } \\ [.1in]
            {\small \bf CC } \\
            {\small There were 16 friends playing a video game online when 7 players quit. If each player left had 8 lives, how many lives did they have total? } \\ [.05in]
            {\small \em Translation produced: } \\
            {\small 8 16 7 - * } \\ [.1in]
            {\small \bf IL } \\
            {\small Lisa flew 256 miles at 32 miles per hour. How long did Lisa fly? } \\ [.05in]
            {\small \em Translation produced: } \\
            {\small 256 32 $/$ } \\ [.1in]
            {\small \bf MAWPS } \\
            {\small Debby's class is going on a field trip to the zoo. If each van can hold 4 people and there are 2 students and 6 adults going, how many vans will they need? } \\ [.05in]
            {\small \em Translation produced: } \\
            {\small 2 6 + 4 $/$ } \\
            \hline
            \end{tabular}
        \end{figure}
        \begin{center}
            \begin{figure}[!ht]
            \caption{Number identification errors.}
            \label{figure:replacementErrors}
            \centering
            \begin{tabular}{p{0.9\linewidth}}
            \hline
            {\small \bf Question (MAWPS) } \\
            {\small Melanie is selling 4 gumballs for eight cents each. How much money can Melanie get from selling the gumballs? } \\ [.05in]
            {\small \bf Correct Translation (Infix) } \\
            {\small 4 * 8 } \\ [.05in]
            {\small \bf Hypothesized Translation } \\
            {\small 4 + $\langle n \rangle$ } \\
            \hline
            \end{tabular}
            \end{figure}
        \end{center}

        Our pre-training was unsuccessful in improving accuracy, even when applied to networks larger than those reported.
        We may need to use more inclusive language, or pre-train on very math specific texts to be successful.
        Our results support our thesis of infix limitation.

    \paragraph{Error Analysis}
        Our system, while performing above standard, could still benefit from some improvements.
        One issue originates from the algorithmic pre-processing of our questions and expressions.
        In Figure \ref{figure:replacementErrors} we show an example of one such issue.
        The excerpt comes from a type 3 non-pre-trained Transformer test.
        The example shows an overlooked identifier, $\langle n1 \rangle$.
        The issue is attributed to the identifier algorithm only considering numbers in the problem.
        Observe in the question that the word ``eight" is the number we expect to relate to $\langle n2 \rangle$.
        Our identifying algorithm could be improved by considering such number words and performing conversion to a numerical value.
        If our algorithm performed as expected, the identifier $\langle n1 \rangle$ relates with 4 (the first occurring number in the question) and $\langle n2 \rangle$ with 8 (the converted number word appearing second in the question).
        The overall translation was incorrect whether or not our algorithm was successful, but it is essential to analyze problems like these that may result in future improvements.
        Had all questions been tagged correctly, our performance would have likely improved.

    \section{Conclusions and Future Work}
        In this paper, we have shown that the use of Transformer networks improves automatic math word problem-solving.
        We have also shown that the use of postfix target expressions performs better than the other two expression formats.
        Our improvements are well-motivated but straightforward and easy to use, demonstrating that the well-acclaimed Transformer architecture for language processing can handle MWPs well, obviating the need to build specialized neural architectures for this task.

        Extensive pre-training over much larger corpora of language has extended the capabilities of many neural approaches.
        For example, networks like BERT \cite{devlin2018bert}, trained extensively on data from Wikipedia, perform relatively better in many tasks.
        Pre-training on a much larger corpus remains an extension we would like to try.

        We want to work with more complex MWP datasets.
        Our datasets contain basic arithmetic expressions of +, -, * and /, and only up to 3 of them.
        For example, datasets such as Dolphin18k \cite{huang2016well}, consisting of web-answered questions from Yahoo! Answers, require a wider variety of arithmetic operators to be understood by the system.

        We have noticed that the presence of irrelevant numbers in the sentences for MWPs limits our performance.
        We can think of such numbers as a sort of adversarial threat to an MWP solver that stress-test it.
        It may be interesting to explore how to keep a network's performance high, even in such cases.

        With a hope to further advance this area of research and heighten interests, all of the code and data used is available on GitHub.\footnote{https://github.com/kadengriffith/MWP-Automatic-Solver}

    \section*{Acknowledgement}
        The National Science Foundation supports the work reported in this paper under Grant No. 1659788.
        Any opinions, findings any conclusions or recommendations expressed in this work are those of the author(s) and do not necessarily reflect the views of the National Science Foundation.

    \balance
    \renewcommand{\baselinestretch}{.65}
    \bibliographystyle{IEEEtran}
    \bibliography{Griffith_Kalita_CSCI4293}

\begin{thebibliography}{10}
\providecommand{\url}[1]{#1}
\csname url@samestyle\endcsname
\providecommand{\newblock}{\relax}
\providecommand{\bibinfo}[2]{#2}
\providecommand{\BIBentrySTDinterwordspacing}{\spaceskip=0pt\relax}
\providecommand{\BIBentryALTinterwordstretchfactor}{4}
\providecommand{\BIBentryALTinterwordspacing}{\spaceskip=\fontdimen2\font plus
\BIBentryALTinterwordstretchfactor\fontdimen3\font minus
  \fontdimen4\font\relax}
\providecommand{\BIBforeignlanguage}[2]{{%
\expandafter\ifx\csname l@#1\endcsname\relax
\typeout{** WARNING: IEEEtran.bst: No hyphenation pattern has been}%
\typeout{** loaded for the language `#1'. Using the pattern for}%
\typeout{** the default language instead.}%
\else
\language=\csname l@#1\endcsname
\fi
#2}}
\providecommand{\BIBdecl}{\relax}
\BIBdecl

\bibitem{vaswani2017attention}
A.~Vaswani, N.~Shazeer, N.~Parmar, J.~Uszkoreit, L.~Jones, A.~N. Gomez,
  {\L}.~Kaiser, and I.~Polosukhin, ``Attention is all you need,'' in
  \emph{Advances in neural information processing systems}, 2017, pp.
  5998--6008.

\bibitem{wang2018mathdqn}
L.~Wang, D.~Zhang, L.~Gao, J.~Song, L.~Guo, and H.~T. Shen, ``Mathdqn: Solving
  arithmetic word problems via deep reinforcement learning,'' in
  \emph{Thirty-Second AAAI Conference on Artificial Intelligence}, 2018.

\bibitem{hosseini2014learning}
M.~J. Hosseini, H.~Hajishirzi, O.~Etzioni, and N.~Kushman, ``Learning to solve
  arithmetic word problems with verb categorization,'' in \emph{Proceedings of
  the 2014 Conference on Empirical Methods in Natural Language Processing
  (EMNLP)}, 2014, pp. 523--533.

\bibitem{kushman2014learning}
N.~Kushman, Y.~Artzi, L.~Zettlemoyer, and R.~Barzilay, ``Learning to
  automatically solve algebra word problems,'' in \emph{Proceedings of the 52nd
  Annual Meeting of the Association for Computational Linguistics (Volume 1:
  Long Papers)}, 2014, pp. 271--281.

\bibitem{roy2015reasoning}
S.~Roy, T.~Vieira, and D.~Roth, ``Reasoning about quantities in natural
  language,'' \emph{Transactions of the Association for Computational
  Linguistics}, vol.~3, pp. 1--13, 2015.

\bibitem{robaidek2018data}
B.~Robaidek, R.~Koncel-Kedziorski, and H.~Hajishirzi, ``Data-driven methods for
  solving algebra word problems,'' \emph{arXiv preprint arXiv:1804.10718},
  2018.

\bibitem{bobrow1964natural}
D.~G. Bobrow, ``Natural language input for a computer problem solving system,''
  Ph.D. dissertation, Massachusetts Institute Of Technology, 1964.

\bibitem{bakman2007robust}
Y.~Bakman, ``Robust understanding of word problems with extraneous
  information,'' \emph{arXiv preprint math/0701393}, 2007.

\bibitem{liguda2012modeling}
C.~Liguda and T.~Pfeiffer, ``Modeling math word problems with augmented
  semantic networks,'' in \emph{International Conference on Application of
  Natural Language to Information Systems}.\hskip 1em plus 0.5em minus
  0.4em\relax Springer, 2012, pp. 247--252.

\bibitem{shi2015automatically}
S.~Shi, Y.~Wang, C.-Y. Lin, X.~Liu, and Y.~Rui, ``Automatically solving number
  word problems by semantic parsing and reasoning,'' in \emph{Proceedings of
  the 2015 Conference on Empirical Methods in Natural Language Processing},
  2015, pp. 1132--1142.

\bibitem{mitra2016learning}
A.~Mitra and C.~Baral, ``Learning to use formulas to solve simple arithmetic
  problems,'' in \emph{Proceedings of the 54th Annual Meeting of the
  Association for Computational Linguistics (Volume 1: Long Papers)}, 2016, pp.
  2144--2153.

\bibitem{koncel2015parsing}
R.~Koncel-Kedziorski, H.~Hajishirzi, A.~Sabharwal, O.~Etzioni, and S.~D. Ang,
  ``Parsing algebraic word problems into equations,'' \emph{Transactions of the
  Association for Computational Linguistics}, vol.~3, pp. 585--597, 2015.

\bibitem{roy2017unit}
S.~Roy and D.~Roth, ``Unit dependency graph and its application to arithmetic
  word problem solving,'' in \emph{Thirty-First AAAI Conference on Artificial
  Intelligence}, 2017.

\bibitem{sun2019neural}
R.~Sun, Y.~Zhao, Q.~Zhang, K.~Ding, S.~Wang, and C.~Wei, ``A neural semantic
  parser for math problems incorporating multi-sentence information,''
  \emph{ACM Transactions on Asian and Low-Resource Language Information
  Processing (TALLIP)}, vol.~18, no.~4, p.~37, 2019.

\bibitem{rehman2019automatically}
T.~Rehman, S.~Khan, G.-J. Hwang, and M.~A. Abbas, ``Automatically solving
  two-variable linear algebraic word problems using text mining,'' \emph{Expert
  Systems}, vol.~36, no.~2, p. e12358, 2019.

\bibitem{huang-etal-2018-neural}
\BIBentryALTinterwordspacing
D.~Huang, J.~Liu, C.-Y. Lin, and J.~Yin, ``Neural math word problem solver with
  reinforcement learning,'' in \emph{Proceedings of the 27th International
  Conference on Computational Linguistics}.\hskip 1em plus 0.5em minus
  0.4em\relax Santa Fe, New Mexico, USA: Association for Computational
  Linguistics, Aug. 2018, pp. 213--223. [Online]. Available:
  \url{https://www.aclweb.org/anthology/C18-1018}
\BIBentrySTDinterwordspacing

\bibitem{peters2018deep}
M.~Peters, M.~Neumann, M.~Iyyer, M.~Gardner, C.~Clark, K.~Lee, and
  L.~Zettlemoyer, ``Deep contextualized word representations,'' pp. 2227--2237,
  2018.

\bibitem{budzianowski2019hello}
\BIBentryALTinterwordspacing
P.~Budzianowski and I.~Vuli{\'c}, ``Hello, it{'}s {GPT}-2 - how can {I} help
  you? towards the use of pretrained language models for task-oriented dialogue
  systems,'' in \emph{Proceedings of the 3rd Workshop on Neural Generation and
  Translation}.\hskip 1em plus 0.5em minus 0.4em\relax Hong Kong: Association
  for Computational Linguistics, Nov. 2019, pp. 15--22. [Online]. Available:
  \url{https://www.aclweb.org/anthology/D19-5602}
\BIBentrySTDinterwordspacing

\bibitem{devlin2018bert}
J.~D. M.-W.~C. Kenton and L.~K. Toutanova, ``Bert: Pre-training of deep
  bidirectional transformers for language understanding.''

\bibitem{roy2016solving}
S.~Roy and D.~Roth, ``Solving general arithmetic word problems,'' pp.
  1743--1752, 2015.

\bibitem{koncel2016mawps}
R.~Koncel-Kedziorski, S.~Roy, A.~Amini, N.~Kushman, and H.~Hajishirzi, ``Mawps:
  A math word problem repository,'' in \emph{Proceedings of the 2016 Conference
  of the North American Chapter of the Association for Computational
  Linguistics: Human Language Technologies}, 2016, pp. 1152--1157.

\bibitem{radford2019language}
A.~Radford, J.~Wu, R.~Child, D.~Luan, D.~Amodei, and I.~Sutskever, ``Language
  models are unsupervised multitask learners,'' \emph{OpenAI Blog}, vol.~1,
  no.~8, 2019.

\bibitem{maas-EtAl:2011:ACL-HLT2011}
\BIBentryALTinterwordspacing
A.~L. Maas, R.~E. Daly, P.~T. Pham, D.~Huang, A.~Y. Ng, and C.~Potts,
  ``Learning word vectors for sentiment analysis,'' in \emph{Proceedings of the
  49th Annual Meeting of the Association for Computational Linguistics: Human
  Language Technologies}.\hskip 1em plus 0.5em minus 0.4em\relax Portland,
  Oregon, USA: Association for Computational Linguistics, June 2011, pp.
  142--150. [Online]. Available: \url{http://www.aclweb.org/anthology/P11-1015}
\BIBentrySTDinterwordspacing

\bibitem{de2005tutorial}
P.-T. De~Boer, D.~P. Kroese, S.~Mannor, and R.~Y. Rubinstein, ``A tutorial on
  the cross-entropy method,'' \emph{Annals of operations research}, vol. 134,
  no.~1, pp. 19--67, 2005.

\bibitem{papineni2002bleu}
K.~Papineni, S.~Roukos, T.~Ward, and W.-J. Zhu, ``Bleu: A method for automatic
  evaluation of machine translation,'' in \emph{Proceedings of the 40th annual
  meeting on association for computational linguistics}.\hskip 1em plus 0.5em
  minus 0.4em\relax Association for Computational Linguistics, 2002, pp.
  311--318.

\bibitem{huang2016well}
D.~Huang, S.~Shi, C.-Y. Lin, J.~Yin, and W.-Y. Ma, ``How well do computers
  solve math word problems? large-scale dataset construction and evaluation,''
  in \emph{Proceedings of the 54th Annual Meeting of the Association for
  Computational Linguistics (Volume 1: Long Papers)}, 2016, pp. 887--896.

\end{thebibliography}
\end{document}